\documentclass{ws-ijait}
\usepackage{cite}
\usepackage{amsmath,amssymb,amsfonts}
\usepackage{algorithm,algorithmic}
\usepackage{changepage}   
\usepackage{graphicx}
\usepackage{tikz}
\tikzset{
	treenode/.style = {shape=rectangle, rounded corners,
		draw, align=center,
		top color=white, bottom color=blue!20},
	root/.style     = {treenode, font=\Large, bottom color=red!30},
	env/.style      = {treenode, font=\ttfamily\normalsize},
	dummy/.style    = {circle,draw}
}
\usepackage{textcomp}
\usepackage{xcolor}
\begin{document}
\markboth{Yitzhak Spielberg and Amos Azaria}
{Criticality-Based Varying Step-Number Algorithm for Reinforcement Learning}

\title{Criticality-Based Varying Step-Number Algorithm for Reinforcement Learning}

\author{Yitzhak Spielberg}
\address{Data-Science Center and Computer Science Dept.\\
\textit{Ariel University\\
Ariel, Israel \\
yspielb@gmail.com}}

\author{Amos Azaria}
\address{Data-Science Center and Computer Science Dept. \\
Ariel University\\
Ariel, Israel \\
amos.azaria@ariel.ac.il}

\maketitle
\begin{abstract}  
In the context of reinforcement learning we introduce the concept of criticality of a state, which indicates the extent to which the choice of action in that particular state influences the expected return. That is, a state in which the choice of action is more likely to influence the final outcome is considered as more critical than a state in which it is less likely to influence the final outcome.

We formulate a criticality-based varying step number algorithm (CVS) - a flexible step number algorithm that utilizes the criticality function provided by a human, or learned directly from the environment. We test it in three different domains including the Atari Pong environment, Road-Tree environment, and Shooter environment. We demonstrate that CVS is able to outperform popular learning algorithms such as Deep Q-Learning and Monte Carlo.

\end{abstract}


\section{Introduction}
\par
Our decisions are not uniform with relation to the consequences they produce. Some of them can be easily and immediately forgotten, while others have very significant consequences that may influence us for the rest of our lives. ``What should one have for dinner?",  ``Should one invest two extra hours to work on one's project or spend the evening watching a movie?",  ``Which route should one take to work?" These decisions are almost meaningless, since they do not have any enduring influence on a person's life. 
On the other end of this spectrum are questions such as: ``In which country does one want to live?", ``Which profession should one possess?" , ``How much to invest in health?", and ``How to educate one's children?". 
These decisions might influence us personally and as well as our close ones for many years to come and therefore require profound consideration. 
\par
In reinforcement learning an autonomous agent is trained to act in a way that maximizes its expected return in a given environment. During the learning process the agent is situated  in a certain state and is required to choose one particular action from a set of possible actions. Clearly, in some situations, different actions may lead to very similar expected return values, while in other situations, different actions may lead to very different expected returns. In the former case we may say that the situation (or state) that the agent is visiting is not very critical, as it does not matter that much which action the agent will choose. However, the second situation appears to be critical, as an agent failing to take an optimal action may result at a very low outcome. 

In this paper we introduce the concept of criticality. The criticality level of a state indicates how much the choice of the action influences the agent's performance. The concept of criticality is inspired by the intuition that a state in which the choice of action matters should be considered as more critical, than a state in which it doesn't. 

\par
We believe that the concept of criticality is particularly useful in the context of human-aided reinforcement learning, where the learning agent receives criticality information from a human trainer. In such a learning scenario there might be algorithms that use criticality in order to boost the agent's performance. In this paper we present one such learning algorithm: the criticality-based varying step-number algorithm (CVS). CVS might be regarded as an algorithm that is closely related to the class of n-step learning algorithms (with a fixed step-number), such as n-step SARSA and n-step Tree Backup, but with a flexible step-number. By using a flexible step-number, CVS does not suffer from the central problem of fixed step-number algorithms: the problem of choosing an appropriate step-number.

We compare the performance of CVS to other reinforcement learning methods in three different domains. While the first two domains are quite simple, they provide strong motivation for the use of CVS. The final domain is an Atari-based domain, namely the game of Pong. We show that CVS outperforms other baselines in these three domains.
%

\section {The Concept of Criticality in Reinforcement Learning}
\subsection {A Definition of Criticality}
\par
In the context of reinforcement learning the criticality of a state indicates how much the choice of action in that particular state influences the expected return.
We define the criticality of a state as a measure of variability of the expected return with respect to the available actions. The criticality is a value in the range of [0,1], where 0 represents no variability between the expected return of the actions (for example, if there is only a single action, or if
all actions result in the same expected return), and 1 represents high variability between the expected return of the actions (for example when some actions result in a very high expected return, while other actions result in a very low expected return). Variability is related
to variance, such that a variance of 0 in the expected return entails variability of 0 (and thus criticality of 0); while a variance greater
than 0 entails criticality greater than 0. 
\par
The recognition that some states are more critical than others is particularly useful in learning situations that include a teacher and a student. An example of such a learning situation is a driving lesson. If a student driver approaches an
obstacle on the road, her teacher may state to her that she must watch out, without suggesting exactly which action to take (e.g. slowing down, turning the wheel right or left etc.). This warning will motivate the student to pay more attention to the situation and therefore it will be more likely that she will be able to avoid the obstacle. Moreover, even if the car later hits that obstacle, the student will understand that she probably took a wrong action back when the teacher warned her, and this understanding will help her to learn more efficiently. The situation of a driving lesson possesses the characteristics of a human-aided reinforcement learning scenario. The learning agent finds himself in a certain state and needs to choose one action from an array of possible actions. The human teacher informs him about the criticality level of the current state. The learning agent then utilizes the criticality information in order to improve his learning strategy (for example by implementing the CVS algorithm, which will be presented in this paper).

\par
We introduced criticality in a way that portraits it as a human centered concept, in the sense that it is a \textit{person's} estimate of the spread of consequences with respect to the available actions. Therefore, the definition implies that the criticality function (that is, the function that assigns a criticality level to each state of the environment) of a given environment is not unique, but can be any element from a whole class of functions that are loosely defined by the variance of the expected return. Beyond this type of diversity there is another dimension of freedom in the concept of criticality, which comes from the absence of the optimal policy in its definition. Since in many environments a human does not exactly know the optimal policy, any definition of criticality that includes the optimal policy in an explicit manner (for example the variance of the optimal Q-function in a given state with respect to the actions) would not be human-friendly. 

\subsection {Obtaining criticality from a model or from the environment}

So far we have discussed a scenario in which the human trainer provides the criticality level in every state encountered by the learning agent. If the human can implement the criticality measure in a functional form (as we later use in the experiment sections), the workload on the human trainer is reasonable. However, a setting, where the human trainer provides criticality in real time during the learning procedure, might be unfeasible for two reasons. Firstly, a learning procedure that takes long would require a substantial investment of time from the trainer. Secondly, because the effort required for the estimation of the criticality level of one single state accumulates over the complete learning session, the trainer might be exposed to a tremendous workload.
\par
There are multiple approaches towards a solution for this problem. The first one involves the human trainer and a criticality model. In this approach the trainer is being asked to give his criticality estimates on a set of states. 
On the basis of this set a criticality model for the given environment is learned. During the reinforcement learning process, the agent obtains its criticality input from the criticality model. An alternative approach is for the reinforcement learner to obtain the criticality level from the environment directly, without the necessity of a human trainer. Since, according to the definition, criticality is related to the variance of the action-value function with respect to the actions, this variance (possibly normalized, because the criticality needs to be in [0,1]) can be used as an estimate of the criticality.

\subsection {Policy-dependent Criticality}
\par
It may not always be obvious which states should be considered critical and which states should be considered as non-critical. For example, a car driving on a straight road with no traffic may seem as being in a non-critical state. However, a driver that suddenly turns the wheel right (or left), may result in hitting a wall, and action that is likely associated with a negative reward. This could imply that the state was in fact a critical state. 
However, in this example the variance might be low, since most actions such as changing the speed or modestly turning the wheel won't have any meaningful impact. 
Therefore, in certain learning situations, it might be necessary to refine the definition of criticality in order to capture these scenarios.
\par
One option is to multiply each expected return by the probability that the agent will take each action, and then compute the weighted variance (rather than the plain variance). This definition may be closer to what humans view as critical states. It would require transforming the weighted variance to a value between 0 and 1 by some kind of normalization procedure. According to this more sophisticated definition, the criticality is no longer associated only with a state, but is now associated with a policy as well, and may therefore change over time. This is intuitive, since when the agent plays better, different states may seem more critical. For example, for a novice basket-ball player, a position from which a 3-point opportunity exists, seems less critical (because the player is very likely to miss) than for a professional player, who is more likely to score.

\section{The Construction of Criticality Measures in Various Environments}
So far we have defined the concept of criticality in reinforcement learning and we have discussed how it can be refined and expanded in order to guarantee more robustness in various learning situations. We have stated that a central feature of the concept of criticality, the way we envision it, is its human-friendliness.
Therefore we formulated our definition in a manner that leaves multiple degrees of freedom by linking criticality only loosely to both the optimal policy and the variance of the Q-function in a given state. In this section we want to convey to the reader an intuition of the way a criticality measure can be constructed by presenting plausible criticality measures in multiple environments.
\par
The Atari Pong environment consists of two rackets, of which one is the agent and the other is the opponent, a ball, and a playing field. 
The agent receives a reward of +1 when he scores a point, and a reward of -1 when the opponent does. The Pong game has an interesting characteristic: when the ball is moving away from the agent, its actions are irrelevant. Plausible criticality measures can be constructed on the basis of this characteristic. The simplest criticality measure could assign a criticality of zero to each state, in which the ball moves away from the agent and a maximal criticality of 1 to each state, in which the ball moves towards the agent. A slightly more sophisticated criticality measure might use some decreasing function of the distance between the ball and the agent in those states, where the ball is moving towards the agent, since the agent's actions become more critical, as the ball is coming closer to it.

Consider a life-guard agent, which is required to ensure the safety of people bathing in a pool. The agent may perform several actions such as throwing a life ring to different locations at the pool. Clearly, one of the most important tasks for such an agent is to detect which states are critical and which states are not. A critical state would be a state in which a person is having some difficulty to remain above water. Taking no action or throwing a life ring to an incorrect location when a person is drowning, may have catastrophic consequences. However, throwing a life ring to any location, when there is no person requiring help, is likely to result in a very minor penalty.

\par 
Self-driving cars are currently one of the most attention-grabbing applications of artificial intelligence. Since reinforcement learning techniques are instrumental in teaching them to drive autonomously, it might be particularly interesting to discuss the construction of a criticality measure which might make the learning procedure more effective. Obviously, the list of critical traffic situations might become very long, because of the complexity of real-world scenarios, so we will limit our scope and indicate only three major categories of critical states. The first type of critical situations is related to weather conditions. It might include scenarios such as black ice and dense fog. Another category of critical states is related to the complexity of the situation. This category might include such situations as left turns, complex junctions and moments in which the behaviour of nearby vehicles is unclear. The third type of critical situation is related to the traffic density. This category might include areas that are highly populated by pedestrians or playing children.

\section {CVS}
In this section we introduce a practical application of criticality in reinforcement learning: the criticality-based varying step-number algorithm (CVS) - a flexible step-number algorithm that utilizes criticality information, in order to avoid the problem of choosing an appropriate step-number in n-step algorithms (which use a fixed value of $n$), such as n-step SARSA and n-step Tree Backup.

\subsection {The  Relation between Criticality and the Step-number}
All prominent n-step reinforcement learning algorithms, such as n-step SARSA, n-step Expected SARSA and n-step Tree Backup, use a fixed step-number $n$ for bootstrapping, which stays constant both in the course of an episode and during the complete learning process. In our approach we use a varying step-number that is specific to each state encountered during an episode, and we use criticality to determine the appropriate step-number for a given state. 

\par In order to develop some intuition on the way in which criticality could be used to determine an appropriate step-number, we present a simple example. In this example we will work with the n-step SARSA return:
\[
G_{t:t+n}= R_t + \gamma R_{t+1} + ... +\gamma^{n-1}R_{t+n-1}+\gamma^n Q(S_{t+n},A_{t+n})
\]

Let us assume that in our environment most of the states have only one available action, and that there is no randomness in the Markov Decision Process (MDP), that is, a given state action pair determines the next state. Let us further assume that during the learning process the agent encounters some sequence of states-action pairs: 
\[
(S_0,A_0), (S_1,A_1), (S_2,A_2), (S_3,A_3), (S_4,A_4)
\]
 of which only $S_3$ has multiple actions available. In this situation, obviously $S_0,S_1,S_2$ should be assigned a criticality of $0$ (since the agent has no choice, and therefore its action has no influence on the final return value, i.e. the variability of the return is 0) whereas for simplicity we will assign to $S_3$ a criticality of $1$. Clearly, whenever the agent arrives at $S_0$, the next states it visits will always be ($S_1,S_2,S_3$). We would like to determine $n \in \lbrace 1,2,3,4 \rbrace$ should be used for the n-step return $G_{0:n}$ that will serve as  the update target for $Q(S_0,A_0)$. 

Consider the simple 1-step SARSA. This algorithm will update $Q(S_0,A_0)$ towards $G_{0:1}$ and in the next step $Q(S_1,A_1)$ towards $G_{1:2}$. These updates will be repeated in each episode where these states are being visited, so it is easy to see that asymptotically $Q(S_0,A_0)$ will be updated towards $G_{0:2}$. Therefore, there is no benefit from selecting $G_{0:1}$ as the update target for $Q(S_0,A_0)$ versus selecting $G_{0:2}$. Moreover, the selection of $G_{0:2}$ may speed up the convergence. Using the same argument we can conclude that $G_{0:3}$ is a better update target than $G_{0:2}$. However, updating $Q(S_0,A_0)$ towards $G_{0:4}$ may not be the best choice, since the agent may choose a different action at $S_3$, which will lead it to a state that is different from $S_4$. 

\par We now discuss the question of how to construct a criticality-based algorithm that would choose $n=3$ for the update target $G_{0:n}$ for $Q(S_0,A_0)$. One way of doing so is by simply choosing the smallest $n>1$ for which $S_n$ has a criticality above a given threshold (e.g. $0.5$). This algorithm looks appealing due to its simplicity and works well in our simple example. Yet, it has two downsides. First, it is not clear what the threshold should be. Second, it is invariant to the criticality of all the states that precede the $S_n$ which corresponds to the chosen update target $G_{0:n}$ as long as they remain beneath the threshold. This is an important point in a situation where the individual states in a certain domain have a criticality beneath the threshold but the domain of the state space as a whole has a high \emph{cumulative criticality}; that is: the sum of the criticality over states that belong to this domain is high. These considerations motivate an alternative way to use criticality for the choice of a good update target: The CVS algorithm, which we present in the next section.

\subsection{The CVS Algorithm}
We now present a method that on the one hand will choose the appropriate update target $S_3$ in the example from the previous section, and on the other hand will avoid the two downsides of the threshold criticality approach. This method uses the idea of cumulative criticality; It chooses the Q-value of the state $S_n$ with the lowest number $n$ for which  $crit(S_1)+crit(S_2)+...+crit(S_n) \geq 1$ as the update target. The choice of the value $1.0$ as the treshold for the cumulative criticality can be motivated if we consider a binary criticality function that assigns a value of either zero or one to a given state. In that case it would be desirable that the Q-values of the critical states (those, whose criticality is one) would be used as update targets. This method does not suffer from any of the disadvantages of the first method: there is no necessity to determine a threshold and it will produce small step-numbers in more critical domains of the state space. We name this algorithm ``Criticality-based Varying Step-number" (CVS). The update target also depends on the specific algorithm to which CVS is applied: E.g. in the CVS version of Q-Learning it will be $\max_a Q(S_n,a) $; in the CVS version of SARSA it will be $Q(S_n,A_n)$ etc.
\begin{algorithm}
	\caption{\textbf{ The CVS algorithm (SARSA version)}} 
		Given:  criticality function Crit()
		
	\begin{algorithmic}
		\STATE $CritCum(s,a)=0$ for all s,a (cumulative criticality) \\
		WaitList= $\lbrace \rbrace$ (states waiting for update) \\
		pick initial state $S=S_0$ and action $A=A_0$ greedily \\
		while $S \neq Terminal$ \\
		\begin{adjustwidth}{.5cm}{}
			add $(S,A)$ to WaitList \\
			observe $R,S'$ \\
			pick $A'$ greedily \\
			for $(\hat{S},\hat{A})$ in WaitList \\
			\begin{adjustwidth}{.5cm}{}
				
				if $CritCum(\hat{S},\hat{A}) \geq 1$: \\
				\begin{adjustwidth}{.5cm}{}
				 update $Q(\hat{S},\hat{A})$ towards update target $Q(S',A')$ \\
				 delete $(\hat{S},\hat{A})$ from WaitList \\
				 $CrtCum(\hat{S},\hat{A})=0$ \\
				\end{adjustwidth}{}
				else: \\
				\qquad $CritCum(\hat{S},\hat{A})+=Crit(S')$ \\
			\end{adjustwidth}{}
			S,A = S',A'		
			
		\end{adjustwidth}{}
		for $(\hat{S},\hat{A})$ in WaitList \\
		\qquad update $Q(\hat{S},\hat{A})$ \\
		\qquad towards update target $Q(S',A')$
		
	\end{algorithmic}
\end{algorithm}

\section{Evaluation of CVS in the Road-Tree environment}
In this section we introduce the Road-Tree environment, an environment that is particularly appropriate to understand the benefits of CVS.  
We test the algorithm against a number of widely used reinforcement algorithms in order to prove it's efficiency. By default, if not specified otherwise, we do not discount the reward (i.e., $\gamma=1$)
and our initial Q-function is constant over the state-action space. Our default values for epsilon and the learning rate are  $\epsilon=0.1$, and $\alpha=0.1$.
Moreover, in all experiments that are mentioned in this paper we use the Q-Learning version of CVS.

\subsection{The Road-Tree environment}
\begin{figure}
	\centering
	\begin{tikzpicture}
	[
	level distance          = 5em,
	level 1/.style={sibling distance=6em},
	level 2/.style={sibling distance=2em},
	edge from parent/.style = {draw, -latex},
	every node/.style       = {font=\footnotesize},
	sloped
	]
\node [root] {0}
	child { node [dummy] {0}      
		child { node [dummy] {0}
			edge from parent node  {10} }
		child { node [dummy] {7}
			edge from parent node  {15} }
		edge from parent node  {20} }
	child { node [dummy] {1}
		child { node [dummy] {1}
			edge from parent node  {15} }
		child { node [dummy] {1}
			edge from parent node 
			{15}}
		edge from parent node  {10} };
	\end{tikzpicture}
	\caption {Road-Tree example. The number in a critical state (junction or terminal state) represents the reward in that state. The number on an edge is the distance between the corresponding nodes.}
	\label{tree1}
\end{figure}
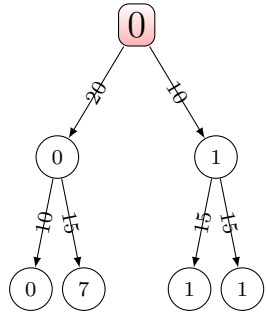

In order to test CVS, we construct a plain environment, named Road-Tree, which has a natural criticality function corresponding to it. Road-Tree has a tree-like structure. The agent starts at the root and always moves in one direction--downward. There are two types of states. In a simple state there is only one possible action. In a junction state the agent needs to choose between multiple roads. The reward upon stepping onto a simple state is always zero. The reward is nonzero only upon reaching a junction or a terminal state. Moreover the reward may vary across junctions and terminal states. Figure \ref{tree1} illustrates a simple Road-Tree environment. The numbers in the junctions represent the rewards. The numbers on the edges show the distance between the two corresponding junctions, which is the number of simple states between them (a distance of $n$ indicates $n-1$ simple states).

The very natural criticality function that we are going to use in the Road-Tree environment assigns zero to a simple state and one to a junction or terminal state.\\

\subsection{CVS vs. Q-Learning, Q($\lambda$) and Monte Carlo}
\begin{figure}
	\centering
	\includegraphics[width=4.6in]{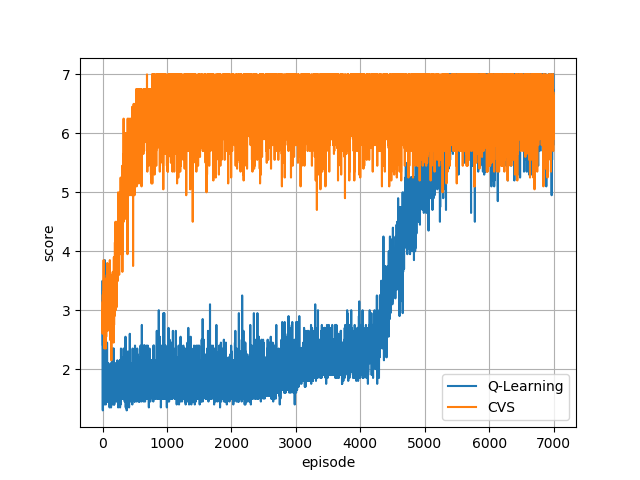}\\
	\caption {CVS against Q-Learning in a 2-level  Road-Tree environment (Figure \ref{tree1}). Average scores over 20 runs. CVS converges to the optimal policy after about 1000 episodes, while Q-Learning requires about 6000 episodes. }
	\label{roadtree_qlearn}
\end{figure}

We now compare the performance of CVS against that of Q-Learning in the 2-level Road-Tree from fig. \ref{tree1}. Clearly, the optimal policy is defined by initially going to the left and then to the right, ending up at the terminal state that has a reward of $r=7$.  
In Q-Learning, due to the relatively big distance between the intermediate junction that has a reward of $r=0$ and the optimal terminal state, the optimal reward ($r=7$) will be backpropagated to the intermediate junction very slowly. The other intermediate junction that has a reward of $r=1$ will be much more attractive to the agent and therefore, the agent might remain in that nonoptimal path for a long period of time. Conversely, the CVS agent will backpropagate the optimal reward terminal state to the intermediate junction immediately after the first visit and therefore should quickly converge to the optimal policy. The plot on fig. \ref{roadtree_qlearn} confirms our elaboration. The Q-Learning agent needs about 6000 episodes to converge to the optimal policy; the CVS agent, in contrast, converges after 1000 episodes. 
\par
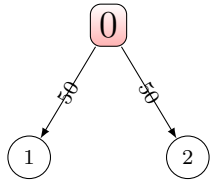
\begin{figure}
	\centering
	\begin{tikzpicture}
	[
	level distance          = 5em,
	level 1/.style={sibling distance=6em},
	edge from parent/.style = {draw, -latex},
	every node/.style       = {font=\footnotesize},
	sloped
	]
	\node [root] {0}
	child { node [dummy] {1}      
		edge from parent node  {50} }
	child { node [dummy] {2}
		edge from parent node  {50} };
	\end{tikzpicture}
	\caption {Plain Road-Tree Environment with only two roads. Since both roads have the same length it is not obvious that CVS will outperform Q($\lambda$). }
	\label{tree2}
\end{figure}

\begin{figure}
	\centering
	\includegraphics[width=4.6in]{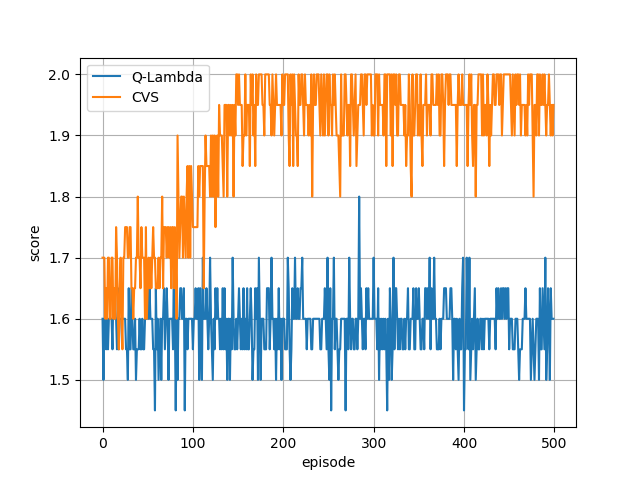}\\
	\caption {CVS performance against Q($\lambda$) in the plain Road-Tree environment fig. \ref{tree2}. Average scores over 20 runs. Q($\lambda$) struggles to make any progress; in contrast, CVS takes about 200 episodes to converge to the optimal policy. }
	\label{roadtree_qlmb}
\end{figure}

 Next we test CVS against Watkin's Q($\lambda$), which is one of the popular algorithms in reinforcement learning (we set $\lambda=0.9$). We perform this evaluation in a very simple Road-Tree environment that contains only two roads that have the same length (see Figure \ref{tree2}). If the optimal road was much longer than the nonoptimal one it would be obvious that CVS would outperform Q($\lambda$),  because of considerations that are very similar to the Q-learning scenario. 
Plot \ref{roadtree_qlmb} shows that even in this more challenging scenario CVS learns faster than Q($\lambda$). It can be seen that Q($\lambda$) struggles to make any progress; in contrast, CVS takes about 200 episodes to converge to the optimal policy.  

\par
\begin{figure}
	\centering
	\begin{tikzpicture}
	[
	level distance          = 5em,
	level 1/.style={sibling distance=6em},
	level 2/.style={sibling distance=2em},
	edge from parent/.style = {draw, -latex},
	every node/.style       = {font=\footnotesize},
	sloped
	]
	\node [root] {0}
	child { node [dummy] {0}      
		child { node [dummy] {0}
			edge from parent node  {} }
		child { node [dummy] {1}
			edge from parent node  {} }
		edge from parent node  {} }
	child { node [dummy] {1}
		child { node [dummy] {-2}
			edge from parent node  {} }
		child { node [env] {...}
			edge from parent node  {} }
		child { node [dummy] {-2}
			edge from parent node  {} }
		child { node [dummy] {1}
			edge from parent node 
			{10}
		}
		edge from parent node  {10} };
	\end{tikzpicture}
	\caption {Road-Tree Environment with two levels and a large number of branches. The terminal state of the optimal trajectory is ``hidden" among 19 suboptimal siblings. }
	\label{tree3}
\end{figure}
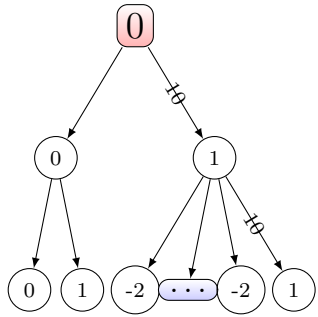

\begin{figure}
	\centering
	\includegraphics[width=4.6in]{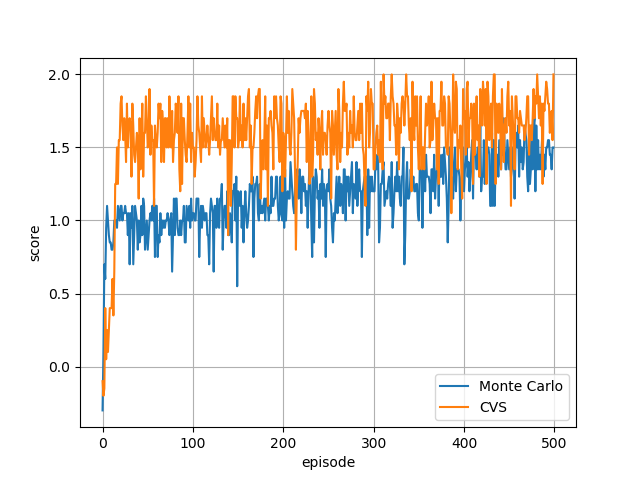}\\
	\caption {CVS vs. Monte Carlo in the Road-Tree environment from fig. \ref{tree3}. Average returns over 20 runs. Monte Carlo requires more than 500 episodes to converge to the optimal policy, CVS only 30. }
	\label{mc_vs_cvs}
\end{figure}

 In the previous example, in which CVS outperformed Q($\lambda$), the algorithm functioned exactly the same way as Monte Carlo (MC) would, by choosing the Monte Carlo return as the update target for each of the previous states. This observation immediately raises the question, whether we can construct an example where CVS would outperform Monte Carlo. 
Figure \ref{tree3}) presents a 3-level Road-Tree with two junctions on the second level and a multitude of terminal states. 
The terminal state of the optimal trajectory is hidden among $99$ siblings, which all have a very bad reward. Since the first visit of the right branch will probably end up at one of these bad siblings, the negative return of the trajectory will be backpropagated to the root immediately and therefore a Monte Carlo agent will 
avoid the right branch, which in fact is the optimal one. In contrast to Monte Carlo, the CVS agent will choose one of the n-step return one of the intermediate junctions,$G_{root:intermjunc}$, as the update target for the root and therefore will not lose its interest in the right branch so quickly. As a consequence, it is much more likely that it will require fewer episodes to discover the optimal trajectory. Indeed, the experiment confirms our intuition. From figure \ref{mc_vs_cvs} we can imply that, as expected, the Monte Carlo agent visits the left junction most of the time and, as a consequence, fails to identify the optimal policy. We can also infer from the plot that, in contrast to the Monte Carlo agent, the CVS agent visits the right junction much more frequently. The plot shows that the optimal trajectory was visited for the first time after about 30 episodes and from there on the CVS agent stayed with it most of the time.

\section{CVS vs. Q-Learning in the Shooter environment}
In this section we describe the performance of CVS versus Q-Learning in a different environment: the Shooter environment. Just like the Road-Tree environment, the Shooter environment can be naturally associated with a simple criticality measure. 

\subsection{The Shooter environment}
The Shooter environment is located on a rectangular playing field of 10x20 (width x length) cells. This playing field contains multiple objects: a gun, which is located in the first column and whose random position may change from game to game; a bullet, which initially is located at the gun's position; and a moving target, which is located in the last column. Each of these objects occupies exactly one cell. Furthermore there exists an obstacle of a size of 3 cells in the 8th column. 
At the beginning of the game the target has a random position in the last column of the field and a random direction of movement, which can be either up or down. In every step the target moves by exactly one cell inside the last column. The direction of the movement is inherited from the previous step with the exception of the case when it hits the wall; in that case the direction is simply being reflected. The agent controls the gun. At any given state of the game the agent can choose one of four actions: Either not shoot at all or shoot in one of the three possible directions - diagonally up, diagonally down or horizontally. The three shooting actions shoot a bullet only if the agent has a bullet to shoot, otherwise these actions are equivalent to doing nothing. 
At any given step the bullet will move by one cell in the direction it was shot; when hitting a wall it's vertical direction is being reflected; if it hits the obstacle the game is terminated with a reward of -1; in the case it reaches the last column, the game is terminated with a reward of +1, if it hits the target or -1, if it does not hit it. 
\begin{figure}
	\centering
	\includegraphics[width=3.0in]{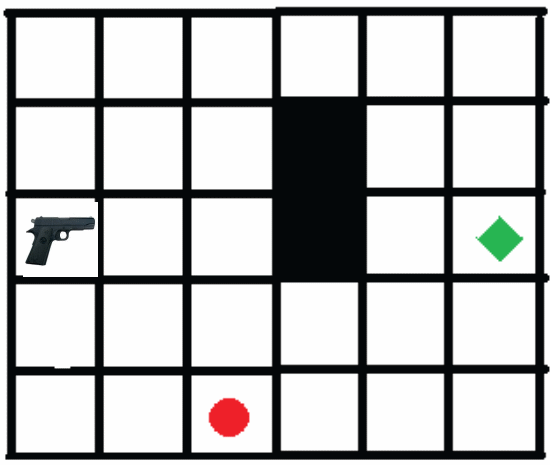}\\
	\caption {Shooter environment. The gun represents the agent's location, the red circle is the bullet, and green diamond represents the target, and the blacked-out squares represent an obstacle. The target is moving, the obstacle isn't. 
    Field size is not the same as in the actual environment. 
    }
	\label{Shooter_env}
\end{figure}
\par There exists a rather natural criticality measure for the Shooter environment. The agent's actions are relevant only before the shot. Moreover before the shot any state can be considered as equally critical. Therefore the most obvious criticality will be binary. It will assign a criticality of 1.0 to any state in which the shot did not take place yet; and a criticality of 0.0 to any state that occurs after the shooting.

\subsection{The performance of CVS vs. Q-Learning}
\begin{figure}
	\centering
	\includegraphics[width=4.6in]{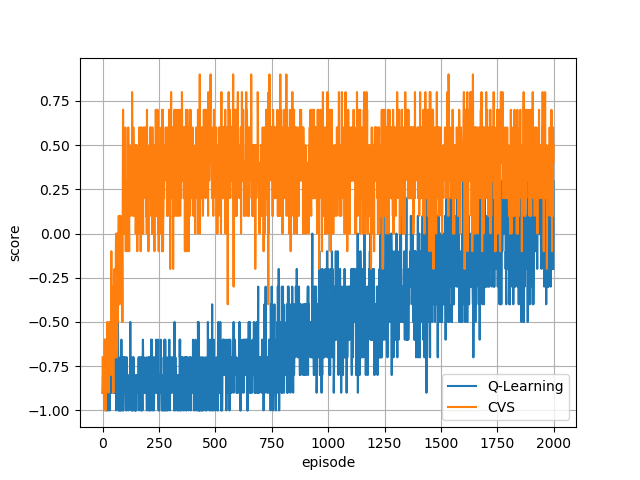}\\
	\caption {A comparison between the performance of CVS against Q-Learning in the Shooter environment. Average scores over 20 runs. Q-Learning requires more than 1500 episodes to reach machine level (score =0.0), CVS only 100 episodes.  }
	\label{plot_avs_shooter}
\end{figure}

In order to compare CVS to Q-Learning, we implemented a tabular Q-Learning agent and a tabular CVS agent. For both agents, we initiated the Q-function to a value of $Q(s)=0$ at every state. The exploration parameter $\epsilon$ was set to a value of 0.1 and remained constant throughout the learning process. The performance of both agents, which was monitored by averaging the scores over 20 runs, is plotted in fig. \ref{plot_avs_shooter}. As depicted in the plot, CVS clearly outperforms Q-Learning. It takes the Q-Learning agent about 1500 episodes to reach an average score of 0.0. Conversely, the CVS agent reaches an average score of 0.0 already after about 100 episodes, and after 200 episodes it converges to a performance level of 0.25. 

\section {Evaluation of CVS in the Atari Pong Environment}
\subsection{The Atari Pong Environment}
The Atari Pong environment consists of two rackets (the agent and the opponent), a ball, and a playing field which has a size of 80x80 pixels. 
The movements of each racket are defined by the three primitive actions (up, down, stay) which either move the racket by several pixels in the corresponding direction or let it remain at the same position. If the racket is located at the wall, and therefore is not able to move in one of the two directions, executing this action is equivalent to staying at the same position. In addition the agent's actions are subject to two noise sources. Firstly the agent will execute the desired action only with a probability p=0.75 and will repeat the previous action with the probability 1-p. Secondly the same action will be executed for $k$ times, where k is being chosen uniformly from the values ${2,3,4}$.  The ball can move in various angles either towards the agent or towards the opponent. If the ball hits either a wall or a racket its direction of movement is reflected. Each game starts with a score of zero and finishes when either the agent or the opponent reaches a score of 21. The agent receives a reward of +1 when it scores, and a reward of -1 when the opponent scores. The initial position of the ball is at the center of the field and the initial direction is always towards the agent. 
\subsection{The DDQN and Monte Carlo algorithms }
In our experiments CVS competes against two algorithms that are located on the extreme ends of the n-step algorithm spectrum: the DDQN algorithm (double DQN) \cite{Hass} which corresponds to $n=1$ and the Monte Carlo algorithm (since, similarly to DDQN, it uses a neural net for the Q-function it can be regarded as a ``deep" Monte Carlo algorithm) which corresponds to $n=\infty$. The main benefit of DDQN over plain DQN is that the second neural net (the target network), which the agent utilizes for action choice, improves the stability of the learning procedure. Similarly to DDQN, we use a target network for action choice in our Monte Carlo implementation too. The strategy to approach the exploration vs. exploitation challenge consists of three learning periods: the first 2000 games are an ``exploration-only period"; afterwards we perform a linear decay of the exploration parameter $\epsilon$ which starts at the value $1.0$ and is finally being decreased to the value of $\epsilon_{fin}=0.1$ by the 5000th game. In the final learning period $\epsilon=\epsilon_{fin}$ is constant. Our learning rate is $\alpha=0.0001$ and our reward decay parameter is $\gamma=0.99$. Our neural net takes the 80x80 image as the input and has an output layer whose size equals the amount of possible actions ( in our case three). It has a compact architecture with only two hidden layers: one convolutional and one fully connected layer. The exact structure is [(Conv,32),(FC,256)].

\subsection{The Implementation of CVS in the Deep-Q-Learning Scenario}
Our implementation of CVS for Deep-Q-Learning is basically a slight variation of the DDQN algorithm. This variation is located in the experience buffer, which is a collection of the agent's previous experiences. Each experience in this buffer consists of two entries: the visited state and the update target. In the DDQN algorithm the update target for a state is always the one-step return. In the implementation of CVS, however, the update target is chosen according to the CVS algorithm.

\subsection{The Choice of the Criticality Function}
We tested two CVS agents. The first agent uses a linear criticality function.  This function is given by a ratio which includes the field length and the distance between the agent's baseline and the ball using the following formula:
\[
crit(s)=1- \frac{dist(ball \ to \ agent's \ baseline)-1}{field \ length -1}
\]
When the ball moves towards the agent, this criticality function takes its minimal value 0 when the ball is at the opponent's racket and its maximal value of 1 when it is one step away from the agent's baseline. When the ball moves away from the agent the criticality is set to 0.
\par
The second CVS agent learns criticality from the environment. His criticality estimate is based on the variance of the Q-function with respect to the actions 
\[
crit(s)= \frac{var_a Q(s,a)}{\max (all \ encountered \ variances) }
\]

\subsection{Atari Pong Environment Results} 
\begin{figure}
	\centering
	\includegraphics[width=4.6in]{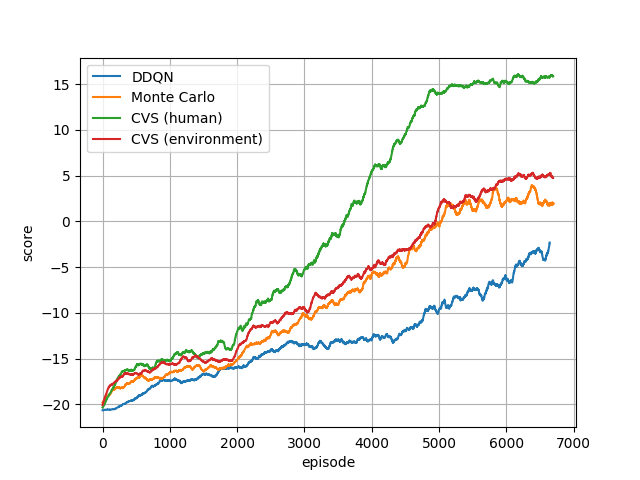}\\
	\caption {CVS vs DDQN and Monte Carlo Atari Pong environment: average scores (5 simulations). The CVS(human) agent works with a criticality function that was designed by a human. The CVS(environment) agent learns the criticality from the environment. The CVS(human) agent clearly outperforms both competitors, whereas the CVS(environment) agent performs very similar to the Monte Carlo agent.}
	\label{Pong_CVS}
\end{figure}

 We plotted the learning performances of four agents: the two CVS agents, the DDQN agent and the Monte Carlo agent. The plot shows scores that were averages over 5 simulations. In order to make the curves smoother, we processed the average scores with a running mean of window size 100. The results of our experiments are shown in figure \ref{Pong_CVS}. One important observation is that the performance boost of CVS(human) in comparison to DDQN is clearly recognizable. The CVS(human) agent after the first 1000 games has only a small lead against the DDQN agent; by game 2500 the lead becomes significant. After about 3500 episodes the CVS(human) agent reaches machine level performance which is about twice as fast as the DDQN agent. The Monte Carlo agent performs better than the DDQN agent as well, although not as good as the CVS(human) agent. While the CVS(environment) agent's performance level seems better than DDQN, it is only slightly better than the Monte Carlo agent, and does not perform as well as the CVS(human) agent, in which the criticality is being provided by the human teacher. 

An analysis of the criticality values that were obtained from the environment showed that, as expected, states in which the ball moved away from the agent, received lower criticality than those where the ball moved towards the agent. However, the spread in criticality values was smaller than in the criticality function that was used for the first CVS agent. We therefore speculate that a weighted variance approach, which takes into account the current policy of the agent (as mentioned above) might work better than the plain variance approach, and will test that approach in future work. 

\par

\section {Related Work}
\par Reinforcement learning based methods have recently shown great success in many domains, including Atari games \cite{Mnih}, Go \cite{Silver}, and autonomous vehicles \cite{Shalev,Sallab,hartmann2019deep,hartmann2020model}. Human-aided reinforcement learning introduces methods that enable the reinforcement learning agent to take advantage of human knowledge in order to learn more efficiently. Prior work in this relatively new area of research has taken a variety of forms. In the first part of this section we present some of these approaches. In the second part we will focus on past research which is more closely related to criticality and to n-step algorithms.

\par One of the ways in which a reinforcement learning agent can profit from human knowledge is by reward-shaping:  engineering an artificial reward function by synthesizing the human's understanding of the environment with the environment's reward function. Reward shaping techniques are particularly appropriate in sparse reward environments such as environments in which all states with the exception of a few terminal states have a zero reward. One of the pioneering reward shaping approaches \cite{Mataric_94} utilized the human's intrinsic knowledge of the environment.  An alternative reward shaping algorithm is the TAMER framework \cite{TAMER} and (the related Deep TAMER \cite{warnell2018deep} for high-dimensional state spaces) which fits a parametric model of the human reward function using human feedback provided during the interactive learning procedure. 

\par Another viable class of methods involve learning from human demonstration \cite{schaal1997learning,li2017sugilite,li2018teaching}. The Human-Agent Transfer algorithm \cite{Taylor_11} is one example from this class. It combines transfer learning, learning from demonstration and reinforcement learning.  Another interesting representative of this class  synthesizes learning from demonstration and reward shaping \cite{Taylor_15}. 

\par Advice plays an important role in the context of human-agent interaction. Advice may be provided by the agent to the human (e.g. \cite{azaria2015advice,azaria2012giving,azaria2012strategic,rosenfeld2015adaptive}) or be provided by the human to assist the agent in its learning process.
Indeed, advice-based techniques 
are used also in human-aided reinforcement learning. In contrast to reward shaping approaches, these techniques instruct the agent directly by feeding it with human advice. Advice-providing methods can be applied in both value-function based and policy-gradient based learning algorithms \cite{Isbell_13,Judah_10}.

\par We now consider work that is more closely related to our paper. We defined the criticality of a state as a subjective measure of the Q-function's variability with respect to the actions. In our literature research we wanted to know whether somewhat similar concepts have been proposed previously. Since similar concepts can be formulated in many different ways the literature research was rather challenging. We found only one concept which is closely connected to criticality and we can not guarantee that we did not miss any other relevant ideas.
This concept, called ``Importance'', was introduced by \cite{Torrey}. The importance of a state is defined by:  
\[
I(s)= \max_a Q(s,a) - \min_a Q(s,a)
\]
The paper proposes multiple algorithms that determine in which states the agent would ask the human teacher for advice and importance was one of the measures which was utilized for this purpose. The ideas formulated in that paper were extended by \cite{Amir:2016} who suggested that advice should be initiated by both the teacher and the agent.
The concept of importance is certainly similar to criticality, since it also measures the Q-function's sensitivity with respect to the action choice. However,  there are also two significant differences between these two concepts. First, the importance of a given state is defined by the current estimate of the Q-function and therefore will change in the course of the learning, while the criticality of a state will not. Second, in contrast to importance, criticality is a purely subjective estimate, which reflects the teachers view of the environment.

\par After having discussed work that is related to the concept of criticality we mention some of the prior research on a topic that is a central problem in n-step algorithms (since CVS is closely related to n-step algorithms): The bias-variance trade-off in n-step algorithms. All n-step algorithms relate to the bias-variance trade-off, since the update of the Q-function suffers from a large bias if the value of $n$ is small, and from large variance if $n$ is big.  Various techniques have been developed to tackle this challenge.
  
\par De Asis \cite{DeAsisPer} addresses this problem for off-policy n-step TD methods, such as n-step Expected SARSA, via the introduction of so called \textit{control variates}. These special terms have the impact of an expectation correction. Therefore they can be used to decrease the bias of the n-step return.

\par Jiang et al. \cite{JiangDoub} propose an alternative solution for this problem for the prediction task (not the optimal control task). They introduce an unbiased estimator, which corrects the current estimate  of the value function $\hat{V}(S_t)$. This estimator is robust in the sense that it remains unbiased even when the function class for the value function is inappropriate.

\par Richard Sutton et al. \cite{SutEmp} suggest an improvement of TD($\lambda$) that achieves an effective bias reduction for the updates. This beneficial effect is a consequence of specific weights that are being assigned to any given update of the value function. The proposed variant of TD($\lambda$) is particularly useful for off-policy learning, where ordinary TD($\lambda$) suffers from a deficit of stability.

\par Unlike all of the above mentioned approaches our method does not manipulate the updates of the (action) value function a-posteriori; instead of doing this, it chooses the appropriate step-number for the update a-priori. This is done by using the criticality function, which is closely related to the update's variance. Therefore, in a broad sense, we can regard the CVS algorithm as a technique that speeds up the learning by controlling the variance of the updates. A preliminary version of this paper appears in \cite{spielberg2019concept}.

\section{Conclusions and Future Work}
We presented the concept of criticality in reinforcement learning and proposed several definitions for it. In the simplest case criticality depends only on the state. A more sophisticated definition might also take into account the agent's current skill level. We introduced the CVS algorithm and tested it in three different domains including the Atari Pong environment. The CVS agent, using a human-designed criticality function, was able to outperform such prominent competitors as DDQN and Monte Carlo. 
Future work will include the development of methods for obtaining criticality functions from human teachers. We consider several methods; the simplest method is by obtaining criticality levels of different states (from a human teacher) and using machine learning to generalize to other states. We will also consider more general approaches in which we will enable users to convey their complete criticality function (likely in a limited set of domains). We will also consider alternatives methods to CVS for using criticality levels. One such method will use criticality levels to determine the contribution of each state using eligibility traces. That is, rewards will be attributed more to critical states than to non-critical states. Each state (or state action pair), will be associated with a weight identical to the criticality, until the sum of all criticality levels reaches 1. If the sum surpasses 1, the final state (which caused the sum to surpass 1), will receive the remainder.
While such an approach might result in faster convergence in terms of the number of episodes, using eligibility traces may require longer to execute.

\par
So far we presented an application of criticality in the domain of reinforcement learning. Yet, the concept of criticality might also have applications in the context of human learning. For example, consider a learning scenario where the human student (a person who learns to play a game) is being assisted by an artificial intelligence agent. One of the ways the agent might support the student in her learning process is by indicating to her which situations are critical. When the student receives an indication that a certain situation is critical, she might pay more attention to it, and consequently, she is more likely to master the challenging situation. 

We also note that beyond enabling the learning assistant to model the criticality function, the procedure of collecting a training set of critical situations might also improve the performance of the human expert. In many tasks that have a monotonous nature (such as surveillance- and monitoring tasks), the operator's attention rapidly decays. An operator that is being asked to record all critical situations, will be much more likely to maintain her vigilance on a high level throughout the complete duration of the task \cite{Hajaj2016}.  
 
\par

\section*{Acknowledgements}
This research was supported in part by the Ministry of Science, Technology \& Space, Israel.

\bibliographystyle{plain}  
\bibliography{mybib.bib}  

\end{document}